\documentclass[11pt]{article}

\usepackage[preprint]{acl}

\usepackage{times}
\usepackage{latexsym}

\usepackage[T1]{fontenc}

\usepackage[utf8]{inputenc}

\usepackage{microtype}

\usepackage{inconsolata}

\usepackage{graphicx}

\title{EmoLASP\/: Emotion Recognition with\\
Language Models and Answer Set Programming}

\author{Thao Le \and Michael Thielscher \\
        School of Computer Science and Engineering, University of New South Wales, Australia \\ {\tt \{thao.le1, mit\}@unsw.edu.au}}

\usepackage{multirow}
\usepackage{booktabs}
\usepackage{amsmath}
\usepackage{amsfonts}
\usepackage[table]{xcolor}

\usepackage{listings}

\defcitealias{katana2010}{UNSW Sydney, 2010}

\lstdefinestyle{promptbox}{
  basicstyle=\ttfamily\small,
  breaklines=true,
  columns=fullflexible,
  frame=single,
  framesep=6pt,
  rulecolor=\color{black!40},
  backgroundcolor=\color{black!3},
  xleftmargin=4pt,
  xrightmargin=4pt,
  literate={–}{{\textendash}}1 {"}{{\textquotedbl}}1,
}

\begin{document}
\maketitle
\begin{abstract}
Emotion recognition in conversations is increasingly tackled with language models, but these models can be unstable and expensive to fine-tune or to prompt with long dialogue histories. We propose \textbf{EmoLASP}, a framework that combines a language model with declarative reasoning via Answer Set Programming (ASP) to predict VAD scores (Valence-Arousal-Dominance) in conversations. Experiments on a widely used benchmark dataset (IEMOCAP) across six open-source LLMs (3B--120B) and two PLMs (BERT, RoBERTa) show that EmoLASP improves prediction performance compared to using the language model alone, even when the LLMs/PLMs are given no dialogue history in their prompts or input vectors. The gains are largest for prompt-only LLMs, which EmoLASP uses without any fine-tuning. However, for fine-tuned PLMs, the reasoner adds little once dialogue history is available. EmoLASP's LLM pipeline demonstrates the potential advantages of using a reasoning approach to ensure emotion prediction consistency and to reduce both the cost of fine-tuning and the cost of prompting with long dialogue histories.
\end{abstract}

\section{Introduction}
Emotion recognition in conversations (ERC) is the task of identifying the emotional state expressed in each utterance of a dialogue, and it supports various applications including empathetic dialogue systems and mental health support~\cite{chu:toward}. Common approaches to ERC include using pre-trained language models (PLMs) and fine-tuning them on labelled emotion datasets to predict emotions~\cite{park2021dimensional,mendes2023quantifying}. In recent years, large language models (LLMs) have been used to recognise emotions in text~\cite{feng2024affect,liu2024emollms}. However, there are several limitations of previous work that need to be addressed. First, most work on ERC focuses on categorical emotion classification, which is less expressive than dimensional representations such as Valence-Arousal-Dominance (VAD)~\cite{russell1977evidence,russell1980circumplex}. Additionally, while LLMs/PLMs have shown promising results in emotion recognition, their instability and bias~\cite{wake2023bias} mean more research is needed to ensure consistent emotional reasoning in conversations. LLMs are essentially black boxes, which might provide an answer without following logical reasoning such as continuity of an emotional state. Moreover, relying solely on language models can be costly, whether due to fine-tuning on domain-specific datasets or the large number of input tokens required when feeding the entire dialogue history in prompt-based approaches.

In this paper, we present a novel approach combining language models with declarative reasoning for emotion recognition. Declarative reasoning has proven effective in other areas such as logical anomaly detection in Vision-Language Models~\cite{jin:logica}. We adopt Answer Set Programming (ASP) as our reasoning framework with the following advantages. First, ASP allows us to express \textit{non-deterministic choice rules}~\cite{eiter2022neuro}, which consider not only the language model's top prediction but also other possible hypotheses with lower confidence. This is particularly useful when outputs are highly uncertain. Second, and on top of this, \textit{weak constraints} in ASP allow us to encode preferences over VAD hypotheses to select the best VAD hypothesis based on the dialogue context. Since \textit{Emotional Contagion Theory} suggests that a person's emotions can be influenced by the emotions of others in a conversation~\cite{hatfield1993emotional}, these weak constraints favour hypotheses consistent with prior turns. Overall, the crucial contribution of the ASP layer is its \textit{declarative} and \textit{extensible} nature, which allows us to define preferences for choosing between hypotheses by stating explicit rules.
Additional preferences, including those derived from other psychological theories, can be incorporated as new rules rather than requiring modifications to imperative code or retraining of the language model.
Our experiments show that the ASP reasoner can handle historical emotion dynamics and select context-consistent emotional hypotheses even when dialogue history is omitted from the language model's input. This is where our cost saving comes from: when EmoLASP is built on prompt-only LLMs, no fine-tuning is performed at all, and omitting the history further reduces the number of input tokens per turn.

\begin{figure*}[ht!]
  \includegraphics[width=\textwidth]{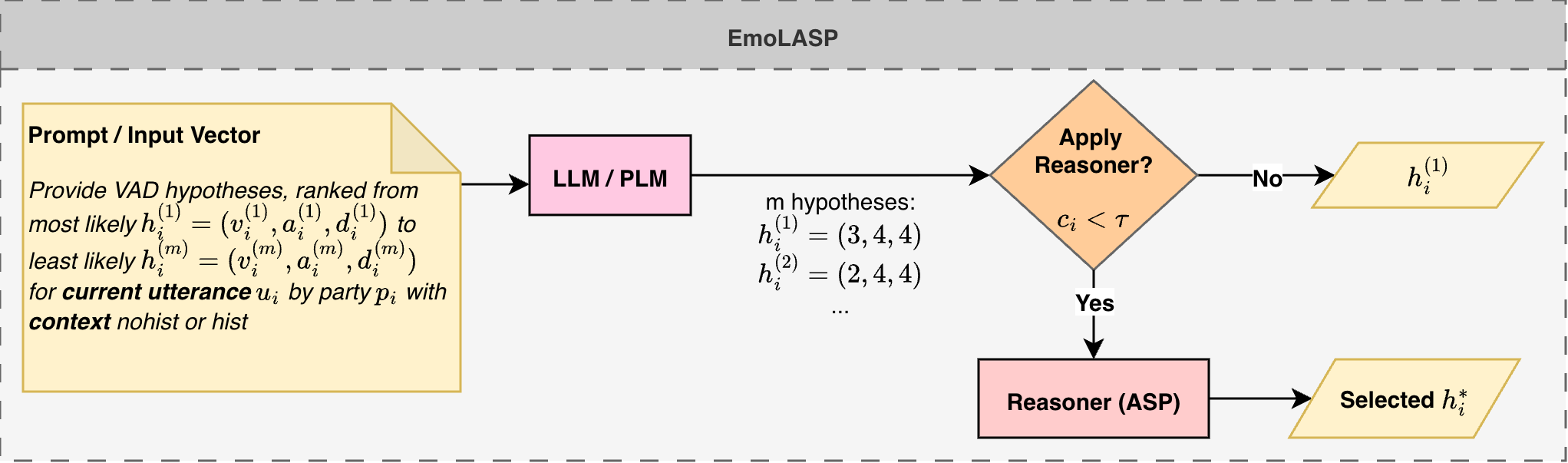}
  \caption{Overview of our EmoLASP framework.}
  \label{fig:emolasp}
\end{figure*}

To our knowledge, this is the first work combining language models and declarative reasoning for ERC. Our contributions are\/: (1) we propose the \textbf{EmoLASP} framework, a novel approach integrating a language model with a reasoner for emotion recognition in conversations; (2) we evaluate it against different open-source language models, including LLMs and PLMs, on a benchmark dataset; (3) we perform qualitative analysis to identify utterances for which LLMs find it most challenging to predict accurate VAD values, even with the help of an ASP reasoner and the presence of dialogue context. Our code is publicly available.\footnote{\url{https://github.com/thaole25/emolasp}}

\section{Method}

Our proposed method has two layers: (1) an LLM/PLM and (2) an answer set program (ASP) implemented in Clingo~\cite{gebser2019multi}. The overall pipeline is as follows: for each utterance, we first use a language model to provide VAD hypotheses. Then, we feed those hypotheses into an ASP reasoner, which performs \textit{aggregation} and \textit{optimisation} to select the best VAD hypothesis for that turn by reasoning about VAD scores of previous turns. Figure~\ref{fig:emolasp} illustrates the overall framework of our method, which we call \textbf{EmoLASP} (\textbf{Emo}tion recognition with \textbf{L}anguage models and \textbf{A}nswer \textbf{S}et \textbf{P}rogramming).

\subsection{Preliminaries}

Given a conversation~$C$ with a sequence of $n$ utterances, \mbox{$ C=\{(u_1,p_1), (u_2,p_2), \ldots, (u_n,p_n)\}$},
where utterance $u_i$ is spoken by party $p_i$, the task of \emph{Emotion Recognition in Conversation} (ERC) is to predict emotional \textbf{VAD} scores for each utterance $u_i$. We write $y_i = (v_i, a_i, d_i)$ for the gold VAD label of $u_i$ and $\hat{y}_i = (\hat{v}_i, \hat{a}_i, \hat{d}_i)$ for our system's prediction. The scores represent a 3-dimensional model of affect, where \textbf{valence} (V) represents the positivity or negativity of an emotion, \textbf{arousal} (A) the intensity or activation level, and \textbf{dominance} (D) the degree of control conveyed by the emotion. VAD scores can be represented as continuous or ordinal values within a specific range.

\subsection{Providing VAD Hypotheses}

Large language models (LLMs) or pre-trained language models (PLMs) provide VAD hypotheses for each utterance. Formally, given an utterance $u_i$, the model produces a set of $m$~VAD hypotheses: $H_i = \{ h_i^{(1)}, h_i^{(2)}, \ldots, h_i^{(m)} \}$, where each $h_i^{(j)} \in \mathbb{R}^3$ is a three-dimensional hypothesis with valence $v_{i}^{(j)}$, arousal $a_{i}^{(j)}$, and dominance $d_{i}^{(j)}$.

\paragraph{LLMs} We prompt an LLM to output $m$ ranked VAD hypotheses for $u_i$, ordered from most to least likely according to the model. Depending on the context condition, the prompt includes either only $u_i$ (\textit{nohist}), or $u_i$ together with all previous turns $\{ u_1, u_2, \ldots, u_{i-1} \}$ (\textit{hist}). The \textit{nohist} condition reduces the number of input tokens, which benefits smaller LLMs with limited context windows.

\paragraph{PLMs} We fine-tune the encoder-based PLMs BERT~\cite{devlin2019bert} and RoBERTa~\cite{liu2019roberta} on the IEMOCAP~\cite{busso2008iemocap} training set with three independent classification heads, one per VAD dimension, each predicting a Likert score in \{1, 2, 3, 4, 5\}.

The \textit{nohist} and \textit{hist} conditions are realised by encoding $u_i$ alone or by prepending the dialogue history with a speaker label before $u_i$. To produce a ranked set of $m$ hypotheses, we apply a softmax to each dimension's logits. Each candidate triple $(v,a,d)$ is scored by the product of its per-dimension probabilities, and the top $m$ unique triples are returned as $H_i$, ranked from most to least likely.

\subsection{Reasoning with Answer Set Programming}
\label{sec:method:asp}
The second layer of our method takes the VAD hypotheses from the language model and applies ASP to optimise emotional dimensions. We first decide whether to invoke ASP using a \textit{confidence threshold}~$\tau$; when invoked, ASP performs two steps: (1)~\textit{aggregation} and (2) \textit{optimisation}.

\paragraph{Confidence Threshold} We apply ASP only when the language model is not highly confident about its predictions. A per-turn confidence score is derived from the language model's own probability distribution (from digit-token logprobs for LLMs, and from the per-dimension softmax product for PLMs). Let $p_i^{(j)}$ be the probability the model assigns to hypothesis $h_i^{(j)}$ for utterance $u_i$, normalised over the $m$ hypotheses. We define the \textit{confidence score} $c_i$ based on Shannon entropy~\cite{shannon1948mathematical} as:
\[
  c_i \;=\; 1 - \frac{E(p_i)}{\ln m}, \quad E(p_i) = -\sum_{j=1}^{m} p_i^{(j)} \ln p_i^{(j)}
\]
Dividing the Shannon entropy $E(p_i)$ by its maximum value $\,\ln m\,$ ensures the score $c_i$ lies in $[0, 1]$ regardless of $m$. We then apply the ASP reasoner when $c_i < \tau$, where threshold $\tau$ is predefined; otherwise, the top-ranked hypothesis is taken as the prediction, $\hat{y}_i = h_i^{(1)}$.

\paragraph{Aggregation} We combine and average the \emph{predicted} VAD scores from the most recent $k$ turns; gold labels $y_j$ are never used at inference time, so there is no label leakage. Formally, let the conversation history before turn $i$ be $C_{i-1} = \{ u_1, u_2, \ldots, u_{i-1} \}$ with corresponding predictions $\hat{Y}_{i-1} = \{ \hat{y}_1, \hat{y}_2, \ldots, \hat{y}_{i-1} \}$. We then combine the~$k$~most recent predicted VAD scores as $ A_i = \frac{1}{k} \sum_{j=i-k}^{i-1} \hat{y}_j$ so that $A_i$ is the aggregated VAD score for turn $i$ based on the most recent $k$ turns.

\paragraph{Optimisation} We use \textit{weak constraints} in ASP to encode objectives for selecting the best VAD hypothesis based on the dialogue context\/:

{\footnotesize
\begin{verbatim}
 :~ pick(N,H), hist_penalty(N,H,P). [P@2, N, H]
 :~ pick(N,H). [H@1, N, H]
\end{verbatim}
}

Intuitively, these constraints operate as follows: Priority \texttt{@2} (first preference) picks the hypothesis~$H$ for turn $N$ that has the smallest distance $P$ to the history aggregate $A_i$. Given a set of hypotheses $H_i$ provided by the language model, let the best hypothesis be $h_i^*$; the distance is then measured using the $L_1$ norm: \mbox{$ h_i^* = \arg\min_{h_i^{(j)} \in H_i} \| h_i^{(j)} - A_i \|_1 $} (the selected VAD hypothesis should be close to the aggregated VAD score, which captures the emotional context from previous turns). When several hypotheses share the minimal distance, priority~\texttt{@1} (second preference) picks the hypothesis with the smallest index $H$, corresponding to the higher-ranked hypothesis from the language model.

\section{Experiment}

\paragraph{Dataset} We use the IEMOCAP dataset~\cite{busso2008iemocap}, which is annotated with VAD scores that serve as labels for evaluation. This dataset contains a total of 115 dialogues across 5 dyadic conversational sessions. Each utterance in the dialogue is annotated with 5-point ordinal ratings (1~to~5) for VAD. The final VAD labels are obtained by averaging the ratings from multiple human annotators. Across 10,039 utterances, annotator counts range from~1 to~4: 8,666 have two annotations, 1,279 have three, 58 have four, and only 36 have just one.

\paragraph{Metrics} There are two metrics used to evaluate the performance. The first is Mean Absolute Error (MAE), which quantifies the average absolute difference between model predictions and the average of human ratings (treated as continuous values). The second is Krippendorff's $\alpha$, which measures how well the model agrees with the human annotators on the original ordinal ratings. To compute~$\alpha$, we add the model's prediction as one extra rater alongside the per-utterance human annotators. We also report Krippendorff's $\alpha$ among only the human annotators as a reference point for human-level agreement.

\begin{table*}[ht!]
\definecolor{hlcol}{gray}{0.86}
\centering
\resizebox{\linewidth}{!}{
{\setlength{\arrayrulewidth}{1.5pt}
\begin{tabular}{l@{\hspace{2pt}}lc>{\columncolor{hlcol}}c>{\columncolor{hlcol}}cc|c>{\columncolor{hlcol}}c>{\columncolor{hlcol}}cc}
\toprule
 &  & \multicolumn{4}{c|}{\textbf{MAE} (lower is better)} & \multicolumn{4}{c}{\textbf{Krippendorff's $\alpha$} (higher is better)} \\
\cmidrule(lr){3-6} \cmidrule(lr){7-10}
 &  & \multicolumn{2}{c}{\textbf{No Reasoner}} & \multicolumn{2}{c}{\textbf{(Ours) EmoLASP}} & \multicolumn{2}{c}{\textbf{No Reasoner}} & \multicolumn{2}{c}{\textbf{(Ours) EmoLASP}} \\
\cmidrule(lr){3-4} \cmidrule(lr){5-6} \cmidrule(lr){7-8} \cmidrule(lr){9-10}
\textbf{Category} & \textbf{Model} & nohist & hist & nohist & hist & nohist & hist & nohist & hist \\
\midrule
\textit{Humans alone} & & \multicolumn{4}{c|}{---} & \multicolumn{4}{c}{$\bar{\alpha}$ = $0.39 \pm 0.03$} \\
\midrule
\multirow{2}{*}{$<10$b} & llama3.2:3b & $0.88 \pm 0.03$ & $1.01 \pm 0.02$ & $\boldsymbol{0.83 \pm 0.02}$ & $0.95 \pm 0.01$ & $\boldsymbol{0.21 \pm 0.02}$ & $0.18 \pm 0.02$ & $\boldsymbol{0.21 \pm 0.02}$ & $0.19 \pm 0.02$ \\
 & gemma3:4b & $0.76 \pm 0.03$$^\dagger$ & $0.84 \pm 0.03$ & $\boldsymbol{0.74 \pm 0.03}$$^\dagger$ & $0.76 \pm 0.04$ & $\boldsymbol{0.26 \pm 0.02}$ & $0.25 \pm 0.02$ & $0.23 \pm 0.02$ & $0.24 \pm 0.02$ \\
\midrule
\multirow{2}{*}{20-30b} & gpt-oss:20b & $0.84 \pm 0.02$ & $0.82 \pm 0.03$ & $0.76 \pm 0.02$ & $\boldsymbol{0.73 \pm 0.02}$ & $0.28 \pm 0.02$ & $\boldsymbol{0.32 \pm 0.02}$$^\dagger$ & $0.28 \pm 0.02$ & $\boldsymbol{0.32 \pm 0.02}$$^\dagger$ \\
 & gemma3:27b & $0.77 \pm 0.01$ & $\boldsymbol{0.75 \pm 0.02}$ & $0.77 \pm 0.01$ & $\boldsymbol{0.75 \pm 0.01}$ & $0.28 \pm 0.02$ & $\boldsymbol{0.30 \pm 0.02}$ & $0.22 \pm 0.02$ & $0.26 \pm 0.03$ \\
\midrule
\multirow{2}{*}{$>60$b} & llama3.3:70b & $0.96 \pm 0.02$ & $0.91 \pm 0.04$ & $\boldsymbol{0.81 \pm 0.02}$ & $\boldsymbol{0.81 \pm 0.03}$ & $0.29 \pm 0.02$$^\dagger$ & $\boldsymbol{0.32 \pm 0.02}$$^\dagger$ & $0.29 \pm 0.02$$^\dagger$ & $\boldsymbol{0.32 \pm 0.02}$$^\dagger$ \\
 & gpt-oss:120b & $0.91 \pm 0.02$ & $0.85 \pm 0.03$ & $0.75 \pm 0.01$ & $\boldsymbol{0.71 \pm 0.02}$$^\dagger$ & $0.27 \pm 0.02$ & $\boldsymbol{0.31 \pm 0.02}$ & $0.27 \pm 0.02$ & $0.30 \pm 0.02$ \\
\midrule
\multirow{2}{*}{PLM} & bert & $0.78 \pm 0.03$ & $0.74 \pm 0.03$ & $0.77 \pm 0.03$ & $\boldsymbol{0.73 \pm 0.03}$ & $0.29 \pm 0.03$$^\dagger$ & $\boldsymbol{0.30 \pm 0.02}$ & $0.29 \pm 0.03$$^\dagger$ & $\boldsymbol{0.30 \pm 0.02}$ \\
 & roberta-large & $0.78 \pm 0.02$ & $0.72 \pm 0.04$$^\dagger$ & $0.77 \pm 0.02$ & $\boldsymbol{0.71 \pm 0.05}$$^\dagger$ & $0.28 \pm 0.02$ & $\boldsymbol{0.32 \pm 0.03}$$^\dagger$ & $0.29 \pm 0.02$$^\dagger$ & $\boldsymbol{0.32 \pm 0.03}$$^\dagger$ \\
\bottomrule
\end{tabular}
}
}
\caption{MAE (left) and Krippendorff's $\alpha$ (right). \textit{Humans alone} is $\alpha$ over only the human annotators. $\dagger$ = best in the same column; \textbf{bold} = best in the same row within its metric block; \textit{gray highlight} = handicap comparison.}
\label{tab:mae_krippendorff_combined}
\end{table*}

\begin{table*}[!ht]
\centering
\resizebox{\linewidth}{!}{
{\setlength{\tabcolsep}{15pt}
\begin{tabular}{llccc|ccc}
\toprule
 &  & \multicolumn{3}{c|}{\textbf{MAE}} & \multicolumn{3}{c}{\textbf{Krippendorff's $\alpha$}} \\
\cmidrule(lr){3-5} \cmidrule(lr){6-8}
\textbf{Category} & \textbf{Model} & \texttt{nohist} & \texttt{hist} & \texttt{handicap} & \texttt{nohist} & \texttt{hist} & \texttt{handicap} \\
\midrule
\multirow{2}{*}{$<10$b} & llama3.2:3b & $\mathbf{-0.049^{***}}$ & $\mathbf{-0.062^{***}}$ & $\mathbf{-0.174^{***}}$ & $-0.002$ & $+0.007$ & $\mathbf{+0.024^{**}}$ \\
 & gemma3:4b & $-0.014$ & $\mathbf{-0.075^{**}}$ & $\mathbf{-0.093^{***}}$ & $-0.029^{**}$ & $-0.010$ & $-0.016$ \\
\midrule
\multirow{2}{*}{20-30b} & gpt-oss:20b & $\mathbf{-0.078^{***}}$ & $\mathbf{-0.091^{***}}$ & $\mathbf{-0.062^{**}}$ & $+0.002$ & $-0.002$ & $-0.036^{***}$ \\
 & gemma3:27b & $+0.002$ & $+0.001$ & $+0.022$ & $-0.061^{***}$ & $-0.047^{***}$ & $-0.079^{***}$ \\
\midrule
\multirow{2}{*}{$>60$b} & llama3.3:70b & $\mathbf{-0.153^{***}}$ & $\mathbf{-0.094^{***}}$ & $\mathbf{-0.097^{***}}$ & $+0.000$ & $+0.000$ & $-0.025^{**}$ \\
 & gpt-oss:120b & $\mathbf{-0.158^{***}}$ & $\mathbf{-0.133^{***}}$ & $\mathbf{-0.096^{***}}$ & $-0.004$ & $-0.004$ & $-0.039^{***}$ \\
\midrule
\multirow{2}{*}{PLM} & bert & $\mathbf{-0.015^{***}}$ & $-0.009$ & $+0.030$ & $\mathbf{+0.006^{*}}$ & $-0.002$ & $-0.012$ \\
 & roberta-large & $\mathbf{-0.014^{**}}$ & $-0.005$ & $+0.051^{*}$ & $\mathbf{+0.004^{*}}$ & $-0.003$ & $-0.029^{*}$ \\
\bottomrule
\end{tabular}
}
}
\caption{Paired $t$-test of \textbf{No Reasoner} vs.\ \textbf{(Ours) EmoLASP} on the IEMOCAP dataset. Each cell reports $\Delta=\overline{\text{EmoLASP}}-\overline{\text{No Reasoner}}$, the mean score difference over the paired units. For MAE a negative $\Delta$ favours EmoLASP; for $\alpha$ a positive $\Delta$ favours EmoLASP. Significance is from a two-sided paired $t$-test: ${}^{*}p<0.05$, ${}^{**}p<0.01$, ${}^{***}p<0.001$. \textbf{Bold} marks a statistically significant improvement by EmoLASP.}
\label{tab:paired_ttest_k2_n3_ge0.25}
\end{table*}

\paragraph{Baselines} Open-source LLMs publicly available on Ollama\footnote{\url{https://ollama.com/}} and two PLMs (BERT, RoBERTa) are used as baselines for comparison. For each language model, there are two context settings: (1)~\textit{nohist}: prompt/input vector has only the current utterance; (2) \textit{hist}: prompt/input vector includes all previous conversation turns and the current utterance. Since there are five different sessions of the IEMOCAP dataset, for PLMs we test on each session and use the remaining four sessions for training. For LLMs, we evaluate on each session separately. Results are reported as the mean and standard deviation across five sessions. Prompts and hyperparameters are detailed in Appendix~\ref{sec:appendix:experiment}.

\section{Results}

\paragraph{RQ1: Can EmoLASP improve ERC performance compared to the LM alone?}

EmoLASP achieves comparable-to-better MAE at substantially lower input-token cost, which is concentrated in the prompt-only LLM pipeline. Table~\ref{tab:paired_ttest_k2_n3_ge0.25} reports paired $t$-tests for every cell of Table~\ref{tab:mae_krippendorff_combined}. Over the 24 MAE comparisons, 16 show a significant improvement of EmoLASP over the No Reasoner baseline, 7 show no significant difference, and only 1 shows a significant decrease. Splitting these by pipeline (LLMs vs. PLMs) makes things clearer. For the six prompt-only LLMs (18 comparisons), 14 show significant improvements and \textit{none} shows a significant decrease; the 4 non-significant cells are all three comparisons for \texttt{gemma3:27b}, plus \texttt{gemma3:4b} under \textit{nohist}. For the two fine-tuned PLMs (6 comparisons), only the two \textit{nohist} cells improve significantly, and the single significant decrease in the whole table is \texttt{roberta-large} under \textit{handicap}.

The \textit{handicap} comparison, highlighted in Table~\ref{tab:mae_krippendorff_combined}, is the key one for the cost claim: it pits the No Reasoner baseline with full dialogue history in the prompt (\textit{hist}) against EmoLASP with no history at all (\textit{nohist}), so any win cannot be attributed to dialogue context. For LLMs, EmoLASP (\textit{nohist}) significantly beats the full-context baseline in 5 of 6 cases, with the remaining case (\texttt{gemma3:27b}) showing no significant difference. For example, MAE drops from 0.91 to 0.81 for \texttt{llama3.3:70b} ($p<0.001$), even though the LLM sees only the current utterance. This matters because EmoLASP (\textit{nohist}) uses fewer input tokens than the No Reasoner baseline (\textit{hist}) and requires no fine-tuning. In contrast, the PLMs behave differently: the reasoner brings no benefit under \textit{handicap} for \texttt{bert} ($p>0.05$) and significantly hurts \texttt{roberta-large} ($0.72 \to 0.77$, $p<0.05$), indicating that fine-tuned PLMs already make effective use of dialogue history in their input. We therefore restrict the cost claim to the prompt-only LLM pipeline, where EmoLASP matches or beats a history-fed baseline without fine-tuning and with fewer input tokens.

The results for Krippendorff's $\alpha$ tell a slightly different story. The human inter-annotator agreement is $0.39$, while the best model $\alpha$ is $0.32$ (e.g., \texttt{gpt-oss:20b}, \texttt{roberta-large} with \textit{hist}). This means that no model reaches human-level agreement yet, but larger LLMs and PLMs come closer than the smaller ones. Across the 24 $\alpha$ comparisons, EmoLASP leaves agreement statistically unchanged in 13, improves it in 3, and decreases it in 8. These decreases are not spread evenly: 3 come from \texttt{gemma3:27b} under all three comparisons, 4 more fall in the \textit{handicap} column, where EmoLASP is denied the dialogue history that the baseline receives, and the last is \texttt{gemma3:4b} under \textit{nohist}. Setting the two \texttt{gemma3} models aside, every matched-context comparison (\textit{nohist} and \textit{hist}) is therefore either unchanged or an improvement, so the MAE gains above do not come at the cost of agreement with human annotators.

\paragraph{RQ2: Which utterances have high VAD prediction errors across all LLMs?}
\label{results:qualitative}
In this section, we identify some examples of utterances that are challenging to predict. A longer list of examples can be found in Appendix~\ref{sec:appendix:qualitative}. A clear pattern appearing across these cases is that while the ground-truth annotations reflect extreme emotions in the VAD space (1 or 5), every model predicts values close to the middle (around 2 to 4). These issues have two main causes. The first involves very short or surface-neutral utterances. One example utterance is \textit{``Hmm''} with human-annotated scores (5, 4, 0.5), meaning the speaker feels very positive, yet every model predicts mid-range valence values from 2 to~3. In another example (\textit{``Are you cold, huh? Do you want to go home?''}), the utterance looks like a caring question, but the gold label (4.50, 4.50, 1.00) reflects strong feeling with very low dominance, which all models miss.

The second cause is figurative or metaphorical language. In one case (\textit{``...I'd like a family. I'd like some kids.''}), the label is (1, 4.5, 5), expressing deep frustration, but models read the words ``family'' and ``kids'' as positive and predict high valence (around 4 to 5). Another example (\textit{``God I feel as though my entire life is going to be spent standing on the beach...waiting for the fish to show up.''}) uses the fish as a metaphor for life-long disappointment, with label (1, 5, 5). Some models (e.g., \texttt{gpt-oss:120b}, \texttt{gpt-oss:20b}) correctly pick up the low valence and high arousal, but they under\-estimate the dominance by predicting 1 instead of 5. ASP does not change this much because LLMs are responsible for generating the VAD hypotheses.

\paragraph{Central-tendency shrinkage} A fair concern is that the MAE gains reflect central-tendency shrinkage, that is, predictions pulled toward the middle of the VAD scale, rather than better emotion recognition. However, three points argue against pure shrinkage. First, the reasoner can only pick among the $m$ hypotheses proposed by the language model, so it cannot regress toward a midpoint that is not already a candidate. It is also invoked only when the model is not confident, leaving confident predictions untouched. Second, if the mechanism were pure smoothing, Krippendorff's $\alpha$ should degrade uniformly. Instead, $\alpha$ is unchanged in 13 of 24 comparisons and improves in 3, and the 8 significant decreases are concentrated in one model (\texttt{gemma3:27b}) and in the \textit{handicap} comparison, where the reasoner is not given the dialogue history that the baseline receives in its prompt. Third, RQ2 shows that every tested model already predicts mid-range values (around 2 to 4) for utterances annotated at the extremes (1 or 5), and ASP cannot help in these cases because the extreme values never appear among the hypotheses it selects from.

\section{Conclusions}
We presented EmoLASP, which adds an ASP reasoner on top of a language model. Our system improves MAE for most prompt-only LLMs, while the fine-tuned PLMs gain little as they already use the dialogue history well. EmoLASP beats a history-fed baseline for 5 of 6 LLMs while seeing only the current utterance, so a small reasoner can replace a long history in the prompt at a lower input-token cost and with no fine-tuning. Agreement with human annotators mostly stays the same, so these gains do not come at the cost of agreement. However, qualitative analysis reveals a limitation of LLM-based approaches: they struggle to accurately interpret extreme emotions in brief utterances or those relying on metaphorical language.

\section*{Limitations}

Our study has several limitations, which concern the data, the framework, and our evaluation. First, regarding the data, we evaluate EmoLASP on a single dataset (IEMOCAP). To the best of our knowledge, this is the only publicly available dataset with human VAD annotations at the dialogue level. While there are other datasets with VAD annotations, such as EmoBank~\cite{buechel2017emobank} or NRC VAD~\cite{mohammad2025nrc}, they are sentence-level or word-level datasets. In the IEMOCAP dataset, human annotators annotate the video data, which includes both audio and visual cues, but our models only have access to the text data. This mismatch may contribute to the disagreement between model predictions and human annotations, and it also means that our models are not fully leveraging the multimodal nature of the data. Also, the number of annotators per utterance varies across utterances, with the majority (8,666 out of 10,039) having only two annotators.

Second, the framework itself has limitations. The current ASP implementation is modest with a choice rule and two weak constraints for optimisation. In future work, we plan to explore more elaborate ASP implementations. The aggregation window $k$, the number of hypotheses $m$, and the confidence threshold $\tau$ are also fixed to $k=2$, $m=3$, and $\tau=0.25$ in our main experiments. Our ablation studies in Appendix~\ref{sec:appendix:ablation} show that the results barely change across a range of these values, and that weighting the history by speaker identity does not help either. However, these sweeps vary one parameter at a time on a single dataset, so the best settings for other datasets or for a joint search may still differ. Furthermore, our analysis in Section~\ref{results:qualitative} shows that the framework still struggles with very short utterances (e.g., \textit{``Hmm''}) and figurative language. Addressing this issue likely requires changes to how hypotheses are generated, which depends on the language models, not just how they are selected by the ASP reasoner.

Third, our evaluation is limited in scope. We currently use fine-tuned PLMs as supervised VAD baselines that are representative of the encoder-based approach and serve as fair points of comparison for the LLM pipeline. Previous ERC models, such as DialogueRNN~\cite{majumder2019dialoguernn}, DialogueGCN~\cite{ghosal2019dialoguegcn} and SKIER~\cite{li2023skier}, predict categorical emotion labels and are not directly comparable on VAD dimensions. Future work could explore the use of these models as baselines by mapping their categorical predictions to VAD scores. Moreover, our PLMs use three independent classification heads over a 5-point Likert scale, which assumes independence across V, A, and D. Finally, all evaluated LLMs are open-source models available through Ollama, and we do not test closed-source models such as GPT-5, so the conclusions may not generalise to all LLMs.

\section*{Ethics Considerations}
\label{sec:appendix:ethics}
We use the IEMOCAP dataset~\cite{busso2008iemocap} for research purposes only, consistent with the terms of its license. Our experiments use only the text transcriptions of utterances and do not process the audio or video recordings. The transcriptions provided by the dataset have been de-identified. In terms of demographics, the corpus consists of acted dyadic dialogues in English, recorded in a single cultural and institutional setting (a US university drama department). The speakers are ten actors (five male, five female) whose demographic characteristics beyond gender are not documented in the original release.

\paragraph{Potential Risks}
Emotion recognition systems carry some risks. First, dimensional VAD predictors can be repurposed for affect-based advertising or surveillance without user consent, even though our motivating applications are empathetic dialogue and mental health support. Second, our qualitative analysis (Section~\ref{results:qualitative}) shows that most evaluated LLMs systematically compress predictions toward the middle of the VAD scale and miss extreme states conveyed through short or figurative utterances. In safety-critical settings such as mental health support, this could cause systems to underestimate distress precisely when accurate recognition matters most, a limitation that EmoLASP does not overcome. Third, IEMOCAP consists of acted English dyadic dialogues from a single cultural setting, which limits generalisation to other populations and languages.

\section*{Acknowledgements}
This research was partially supported by the Australian Research Council (grant \#DP250101822). The experiments were conducted on Katana~\citepalias{katana2010}.

\bibliography{references}

\appendix
\section{AI Assistants in Research or Writing}
We use AI assistants, including Claude and ChatGPT, for coding assistance, grammar checks and stylistic improvements.

\section{Extended Background}
In this section, we provide more details about other related work on applying answer set programming (ASP) to affective computing. \citet{galindo2020friend} introduced an E-friend chatbot for mental well-being that maintains a model of the user's emotion and mental state using logic rules. In their system, psychological theories such as the Ortony, Clore, and Collins (OCC) model~\cite{ortony1988cognitive} are represented as ASP rules. The ASP reasoning module helps the chatbot decide on strategies such as asking questions or suggesting activities to improve the user's emotional state. Another example is the work by~\citet{brannstrom2022emotional} on emotion-aware agent planning, in which the authors formalise emotional dynamics as an action language and translate it to ASP for reasoning about how an agent's actions influence human emotions. Moreover,~\citet{eiter2023logic} also took advantage of weak constraints in ASP, similar to our approach, but in the context of generating counterfactual explanations in the visual question answering domain. This approach uses ASP to provide explanations given an image and a user query; in contrast, we use ASP to select among emotion hypotheses produced by a language model. They nevertheless highlight that ASP can be powerful for ensuring logical consistency and integrating human knowledge about emotions.

Moreover, LM-ASP pipelines have been explored in Natural Language Understanding (NLU) tasks, but to our knowledge, not yet for emotion recognition. For example, a common approach is to use LLMs to translate the problem description and user query into ASP syntax, then apply an ASP solver to perform reasoning tasks~\cite{yang2023coupling,alviano2025integrating}. Another line of work uses LLMs to generate candidate continuations of a story, and ASP for filtering out illogical continuations~\cite{nye2021improving}. In handling dialogues, the LM-ASP combination has been used in the task-oriented chatbot domain~\cite{rajasekharan2023reliable,zeng2024reliable}, where LLMs are used to extract knowledge from the dialogue, and ASP is executed to decide on the next action of the chatbot, such as recommending a restaurant or answering a question. However, those works are restricted to movie and restaurant recommendation domains and have not been applied to emotion understanding. Unlike pipelines that translate the full problem into ASP and let the solver do the heavy reasoning, EmoLASP uses ASP specifically for its choice rules and weak constraints with priorities, which let us encode preferences over LM-produced VAD hypotheses while still leveraging the language model for semantic interpretation.

In the ERC domain, several related works have used purely neural or neuro-symbolic approaches. \citet{majumder2019dialoguernn} applied an RNN to track individual speaker states for emotion classification. \citet{ghosal2019dialoguegcn} presented a graph convolutional network (GCN) to model the inter-speaker dependency. Additionally,~\citet{yu2025beyond} developed a graph neural network to simulate the emotional contagion process~\cite{hatfield1993emotional}. Furthermore, to integrate knowledge beyond the conversation context,~\citet{li2023skier} utilised an external commonsense knowledge base to understand relations between two different concepts and hence infer the emotion class of an utterance from its context. Unlike these neural approaches that predict categorical emotion labels, EmoLASP targets VAD dimensions and uses ASP to enforce contextual consistency over LM-produced hypotheses, without requiring the extensive neural-network training that these methods rely on; such training is outside the scope of our paper.

\section{Experiment Details}
\label{sec:appendix:experiment}
\subsection{Licenses}
Our code is publicly released under the MIT License to allow reproducibility and future research\footnote{\url{https://github.com/thaole25/emolasp}}. The dataset used in this paper, IEMOCAP~\cite{busso2008iemocap}, is shared for research purposes only.

\subsection{Hyperparameters}

Our experiments were conducted on a shared computing infrastructure equipped with NVIDIA V100, A100, and H200 GPUs. We used a single GPU per job. A total of 8 (models) $\times$ 2 (contexts: nohist, hist) $\times$ 5 (dialogue sessions) = 80 jobs were run.

When training PLMs, we set the learning rate to 2e-5, batch size to 16, max context length to 512, and trained for 20 epochs. There are two PLMs: BERT has $\sim 110$ million parameters, and RoBERTa-large has $\sim 355$ million parameters. For each model, we ran 10 jobs (2 contexts $\times$ 5 dialogue sessions). We estimate the total running time to be approximately 10 hours for the two models.

For LLMs, we set the temperature to 0, and the number of VAD hypotheses $m$ to 3. For ASP, we set the confidence threshold $\tau$ to 0.25 and the aggregation window size $k$ to 2. We use six LLMs with sizes ranging from 3 billion to 120 billion parameters. For each model, we ran 10 jobs (2 contexts $\times$ 5 dialogue sessions). The total running time is estimated to be approximately 170 hours for all six LLMs.

\paragraph{Why $\tau=0.25$?} Specifically, with $m=3$, let the highest probability for utterance $u_i$ be $p^*_i=0.7$ and assume the two other hypotheses have the same probability of $0.15$ (highest uncertainty); the corresponding Shannon entropy is $E(p_i) = -0.7 \ln 0.7 - 0.15 \ln 0.15 - 0.15 \ln 0.15 \approx 0.82$. The confidence score is then $c_i = 1 - \frac{E(p_i)}{\ln m} = 1 - \frac{0.82}{\ln 3} \approx 0.25$. Therefore, setting $\tau=0.25$ means that we apply ASP when the language model's best hypothesis has a probability lower than 0.7.
\clearpage
\subsection{Prompt Templates}

We provide the full prompt templates for LLM hypothesis generation in Figures~\ref{fig:prompt-nohist} and \ref{fig:prompt-hist}. The only difference between the two prompts is that the \texttt{hist} prompt includes the dialogue context, while the \texttt{nohist} prompt does not. The 5-point Likert scale for VAD is based on the Self-Assessment Manikin (SAM)~\cite{bradley1994measuring,bradley1999affective}.

\begin{figure}[!ht]
\begin{lstlisting}[style=promptbox]
Analyze the emotion expressed in the following text and provide 3 plausible Valence, Arousal, and Dominance (VAD) hypotheses, ranked from most likely (H1) to least likely (H3).

Each score must be an integer from 1 to 5 (inclusive). Use only whole numbers 1, 2, 3, 4, or 5.

**Valence (V)** - happiness vs unhappiness:
- 1 = Complete unhappiness, annoyance, unsatisfaction, melancholia, or boredom.
- 2 = Mostly negative, low pleasure.
- 3 = Neutral; neither clear happiness nor sadness.
- 4 = Mostly positive, moderate pleasure.
- 5 = Complete happiness, pleasure, satisfaction, contentment, or hope.

**Arousal (A)** - calm vs excited:
- 1 = Complete relaxation, calmness, sluggishness, dullness, sleepiness, or lack of excitement.
- 2 = Mostly calm, low energy.
- 3 = Neutral; neither calm nor excited.
- 4 = Mostly stimulated, moderate energy.
- 5 = Complete stimulation, excitement, frenzy, nervousness, or high arousal.

**Dominance (D)** - controlled vs in-control:
- 1 = Complete feeling of being controlled, influenced, cared-for, awed, submissive, or guided.
- 2 = Mostly submissive, low control.
- 3 = Neutral; neither controlling nor controlled.
- 4 = Mostly in-control, moderate dominance.
- 5 = Complete feeling in-control, influential, important, dominant, autonomous, or controlling.

Text: "<UTTERANCE>"

Return exactly 3 ranked hypotheses, one per line, with H1 the most likely and H3 the least likely:
H1: V=X, A=X, D=X
H2: V=X, A=X, D=X
H3: V=X, A=X, D=X
(X must be one of 1, 2, 3, 4, or 5)
Your output:
\end{lstlisting}
\caption{LLM hypothesis-generation prompt without dialogue context (\texttt{nohist}).}
\label{fig:prompt-nohist}
\end{figure}

\begin{figure}[!ht]
\begin{lstlisting}[style=promptbox]
Analyze the emotion expressed in the current text based on the dialogue context and provide 3 plausible Valence, Arousal, and Dominance (VAD) hypotheses,
ranked from most likely (H1) to least likely (H3).

Each score must be an integer from 1 to 5 (inclusive). Use only whole numbers 1, 2, 3, 4, or 5.

**Valence (V)** - happiness vs unhappiness:
- 1 = Complete unhappiness, annoyance, unsatisfaction, melancholia, or boredom.
- 2 = Mostly negative, low pleasure.
- 3 = Neutral; neither clear happiness nor sadness.
- 4 = Mostly positive, moderate pleasure.
- 5 = Complete happiness, pleasure, satisfaction, contentment, or hope.

**Arousal (A)** - calm vs excited:
- 1 = Complete relaxation, calmness, sluggishness, dullness, sleepiness, or lack of excitement.
- 2 = Mostly calm, low energy.
- 3 = Neutral; neither calm nor excited.
- 4 = Mostly stimulated, moderate energy.
- 5 = Complete stimulation, excitement, frenzy, nervousness, or high arousal.

**Dominance (D)** - controlled vs in-control:
- 1 = Complete feeling of being controlled, influenced, cared-for, awed, submissive, or guided.
- 2 = Mostly submissive, low control.
- 3 = Neutral; neither controlling nor controlled.
- 4 = Mostly in-control, moderate dominance.
- 5 = Complete feeling in-control, influential, important, dominant, autonomous, or controlling.

Dialogue Context:
Person A: <PRIOR UTTERANCE 1>
Person B: <PRIOR UTTERANCE 2>
...

Current Text: "<SPEAKER LABEL>: <UTTERANCE>"

Return exactly 3 ranked hypotheses, one per line, with H1 the most likely and H3 the least likely:
H1: V=X, A=X, D=X
H2: V=X, A=X, D=X
H3: V=X, A=X, D=X
(X must be one of 1, 2, 3, 4, or 5)
Your output:
\end{lstlisting}
\caption{LLM hypothesis-generation prompt with dialogue context (\texttt{hist}). All prior same-dialogue turns are included, each prefixed with its
\texttt{Person A} or \texttt{Person B} speaker label.}
\label{fig:prompt-hist}
\end{figure}


\section{Results}

\begin{table*}[ht!]
\centering
\resizebox{\linewidth}{!}{
\begin{tabular}{llcc|cc}
\toprule
 &  & \multicolumn{2}{c|}{\textbf{MAE} (lower is better)} & \multicolumn{2}{c}{\textbf{Krippendorff's $\alpha$} (higher is better)} \\
\cmidrule(lr){3-4} \cmidrule(lr){5-6}
\textbf{Ablation} & \textbf{Setting} & nohist & hist & nohist & hist \\
\midrule
\multirow{4}{*}{$k$ (history window)} & $k=1$ & $0.79 \pm 0.03$ & $0.78 \pm 0.07$ & $\boldsymbol{0.26 \pm 0.03}$ & $\boldsymbol{0.28 \pm 0.05}$ \\
 & \underline{$k=2$} & $0.78 \pm 0.03$ & $0.77 \pm 0.07$ & $\boldsymbol{0.26 \pm 0.03}$ & $\boldsymbol{0.28 \pm 0.04}$ \\
 & $k=3$ & $\boldsymbol{0.77 \pm 0.03}$ & $0.76 \pm 0.07$ & $\boldsymbol{0.26 \pm 0.04}$ & $0.27 \pm 0.05$ \\
 & $k=5$ & $\boldsymbol{0.77 \pm 0.03}$ & $\boldsymbol{0.75 \pm 0.07}$ & $\boldsymbol{0.26 \pm 0.04}$ & $\boldsymbol{0.28 \pm 0.04}$ \\
\midrule
\multirow{4}{*}{$m$ (\#\,hypotheses)} & \underline{$m=3$} & $0.78 \pm 0.03$ & $0.77 \pm 0.07$ & $\boldsymbol{0.26 \pm 0.03}$ & $\boldsymbol{0.28 \pm 0.04}$ \\
 & $m=5$ & $\boldsymbol{0.76 \pm 0.06}$ & $0.76 \pm 0.09$ & $\boldsymbol{0.26 \pm 0.03}$ & $\boldsymbol{0.28 \pm 0.05}$ \\
 & $m=7$ & $\boldsymbol{0.76 \pm 0.08}$ & $0.76 \pm 0.09$ & $\boldsymbol{0.26 \pm 0.04}$ & $0.27 \pm 0.05$ \\
 & $m=9$ & $\boldsymbol{0.76 \pm 0.08}$ & $\boldsymbol{0.75 \pm 0.09}$ & $0.25 \pm 0.04$ & $0.27 \pm 0.05$ \\
\midrule
\multirow{4}{*}{$\tau$ (confidence gate)} & $\tau=0.1$ & $0.79 \pm 0.03$ & $0.78 \pm 0.08$ & $\boldsymbol{0.26 \pm 0.03}$ & $\boldsymbol{0.28 \pm 0.05}$ \\
 & \underline{$\tau=0.25$} & $0.78 \pm 0.03$ & $0.77 \pm 0.07$ & $\boldsymbol{0.26 \pm 0.03}$ & $\boldsymbol{0.28 \pm 0.04}$ \\
 & $\tau=0.4$ & $\boldsymbol{0.77 \pm 0.03}$ & $0.77 \pm 0.08$ & $\boldsymbol{0.26 \pm 0.04}$ & $\boldsymbol{0.28 \pm 0.04}$ \\
 & $\tau=0.6$ & $\boldsymbol{0.77 \pm 0.04}$ & $\boldsymbol{0.76 \pm 0.08}$ & $\boldsymbol{0.26 \pm 0.03}$ & $\boldsymbol{0.28 \pm 0.04}$ \\
\bottomrule
\end{tabular}
}
\caption{Ablation of the EmoLASP reasoner hyperparameters. Each row holds two of $\{k, m, \tau\}$ at the anchor $(k=2, m=3, \tau=0.25)$ (underlined) and sweeps the third. Cells aggregate across 6 LLMs + 2 PLMs by first averaging across the 5 cross-session splits per model, then reporting mean $\pm$ std across the 8 per-model means. \textbf{Bold} marks the best value within each parameter group and column.}
\label{tab:ablation}
\end{table*}

\begin{table*}[ht!]
\centering
\resizebox{\linewidth}{!}{
\begin{tabular}{lcc|cc}
\toprule
 & \multicolumn{2}{c|}{\textbf{MAE} (lower is better)} & \multicolumn{2}{c}{\textbf{Krippendorff's $\alpha$} (higher is better)} \\
\cmidrule(lr){2-3} \cmidrule(lr){4-5}
\textbf{$w_{\text{self}}{:}w_{\text{other}}$} & nohist & hist & nohist & hist \\
\midrule
$1{:}3$ & $\boldsymbol{0.78 \pm 0.03}$ & $0.78 \pm 0.07$ & $\boldsymbol{0.26 \pm 0.03}$ & $0.27 \pm 0.05$ \\
$1{:}2$ & $\boldsymbol{0.78 \pm 0.03}$ & $0.78 \pm 0.07$ & $\boldsymbol{0.26 \pm 0.04}$ & $0.27 \pm 0.05$ \\
\underline{$1{:}1$ \emph{(neutral)}} & $\boldsymbol{0.78 \pm 0.03}$ & $\boldsymbol{0.77 \pm 0.07}$ & $\boldsymbol{0.26 \pm 0.03}$ & $\boldsymbol{0.28 \pm 0.04}$ \\
$2{:}1$ & $\boldsymbol{0.78 \pm 0.03}$ & $\boldsymbol{0.77 \pm 0.07}$ & $\boldsymbol{0.26 \pm 0.04}$ & $\boldsymbol{0.28 \pm 0.05}$ \\
$3{:}1$ & $\boldsymbol{0.78 \pm 0.03}$ & $\boldsymbol{0.77 \pm 0.07}$ & $\boldsymbol{0.26 \pm 0.03}$ & $\boldsymbol{0.28 \pm 0.04}$ \\
\bottomrule
\end{tabular}
}
\caption{Ablation of the EmoLASP speaker-weighting ratio $w_{\text{self}}{:}w_{\text{other}}$, holding the other reasoner hyperparameters at the anchor $(k=2, m=3, \tau=0.25)$. The $1{:}1$ row (underlined) is the speaker-blind default. Cells aggregate across 6 LLMs + 2 PLMs by first averaging across the 5 cross-session splits per model, then reporting mean $\pm$ std across the 8 per-model means. \textbf{Bold} marks the best value per column.}
\label{tab:ablation_sw}
\end{table*}

\subsection{Speaker-Weighted Aggregation}
\label{sec:appendix:speaker}

The ablation in Table~\ref{tab:ablation_sw} changes only the aggregation step of Section~\ref{sec:method:asp}. In the main method, the aggregate $A_i$ weights the $k$ most recent turns equally. In the speaker-weighted variant, each history turn $j$ is weighted by whether its speaker $p_j$ is the speaker $p_i$ of the target turn:
\[
w_j = \begin{cases} w_{\text{self}} & \text{if } p_j = p_i \\ w_{\text{other}} & \text{if } p_j \neq p_i \end{cases}
\]
and the aggregate becomes the weighted mean
\[
A^{\text{sw}}_i = \frac{1}{W_i}\sum_{j=i-k}^{i-1} w_j\,\hat{y}_j, \quad W_i = \!\!\sum_{j=i-k}^{i-1}\!\! w_j .
\]
Setting $w_{\text{self}} > w_{\text{other}}$ pulls the aggregate toward the speaker's own previous turns, and $w_{\text{other}} > w_{\text{self}}$ toward the interlocutor. The optimisation step is unchanged apart from using $A^{\text{sw}}_i$ in place of $A_i$:
\[
\hat{y}_i = \arg\min_{h_i^{(j)} \in H_i} \lVert h_i^{(j)} - A^{\text{sw}}_i \rVert_1 ,
\]
with ties again broken by the higher-ranked hypothesis.

\subsection{Ablation Studies}
\label{sec:appendix:ablation}

In this section, we present ablation studies to investigate the impact of the hyperparameters: the history window size $k$, the number of hypotheses $m$, and the confidence threshold $\tau$. We also add speaker-weighted aggregation variants to explore whether weighting the history aggregate by speaker identity influences performance. The results are reported in Tables~\ref{tab:ablation} and \ref{tab:ablation_sw}.

In Table~\ref{tab:ablation}, each sweep changes one parameter and keeps the other two at the default setting $(k=2, m=3, \tau=0.25)$. The main finding is that EmoLASP is stable: across all twelve settings, MAE stays between $0.75$ and $0.79$ and Krippendorff's $\alpha$ between $0.25$ and $0.28$. These gaps are much smaller than the standard deviation across models (up to $\pm 0.09$), so no setting is clearly better than another. Looking at a longer history ($k=5$) or asking for more hypotheses ($m=9$) lowers MAE by about $0.02$, but $\alpha$ does not improve and even drops slightly in some cases. In other words, a small gain in average error does not mean better agreement with the human ratings. Changing $\tau$ has the smallest effect of all: raising it from $0.1$ to $0.6$ makes the reasoner run on many more turns, yet the scores barely move, which shows that applying ASP more often does not hurt performance.

Table~\ref{tab:ablation_sw} tests whether the history aggregate should treat the two speakers differently. Instead of averaging the last $k$ turns equally, we give weight $w_{\text{self}}$ to turns from the same speaker and $w_{\text{other}}$ to turns from the interlocutor, as described in Appendix~\ref{sec:appendix:speaker}. The results are almost identical for every ratio: MAE is $0.78$ without history and $0.77$--$0.78$ with history, and $\alpha$ is $0.26$--$0.28$. The $w_{\text{self}} > w_{\text{other}}$ settings ($2{:}1$ and $3{:}1$) match the speaker-blind default, while the $w_{\text{other}} > w_{\text{self}}$ settings ($1{:}2$ and $1{:}3$) are worse on $\alpha$ with history ($0.27$ vs.\ $0.28$). Since no ratio gives a real gain, we keep the simpler speaker-blind $1{:}1$ aggregate.

Overall, these two ablations suggest that EmoLASP is not significantly affected by hyperparameter tuning, which is useful in practice because the reasoner has no trained parameters. We therefore keep the default $(k=2, m=3, \tau=0.25)$ with equal speaker weights.

\subsection{Qualitative Analysis}
\label{sec:appendix:qualitative}
\begin{table*}[!ht]
\centering
\scriptsize
\resizebox{\linewidth}{!}{
\begin{tabular}{rp{3.8cm}p{3.5cm}p{3.2cm}p{3.2cm}c}
\toprule
\textbf{\#} & \textbf{Context Summary} & \textbf{Current Utterance} & \textbf{No Reasoner (hist)} & \textbf{EmoLASP (hist)} & \textbf{Label (V, A, D)} \\
\midrule
1 & \textbf{Summary:} Person B tells Person A that Person C (the father) is still obsessed with missing Larry. Person B insists it's dishonest to pretend Larry is alive and reveals he wants to marry Annie, who is Larry's former girlfriend. However, Person A fears this will devastate Person C and disrupt the family and business.
\newline \textbf{Two prior turns:}
\newline Person B: The business? The business doesn't inspire me. \newline Person A: Must you be inspired? & Person B: Yeah, I'd like one hour a day. If I have to grub for money all day long, I'd like to go home to something beautiful. I'd like a family. I'd like some kids. & gemma3:4b: (4.00, 4.00, 4.00) \newline gemma3:27b: (4.00, 2.00, 3.00) \newline gpt-oss:20b: (5.00, 3.00, 3.00) \newline gpt-oss:120b: (4.00, 2.00, 2.00) \newline llama3.2:3b: (4.00, 3.00, 4.00) \newline llama3.3:70b: (4.00, 3.00, 4.00) & gemma3:4b: (4.00, 4.00, 4.00) \newline gemma3:27b: (4.00, 2.00, 3.00) \newline gpt-oss:20b: (5.00, 3.00, 3.00) \newline gpt-oss:120b: (3.00, 3.00, 2.00) \newline llama3.2:3b: (4.00, 3.00, 3.00) \newline llama3.3:70b: (4.00, 3.00, 4.00) & (1.00, 4.50, 5.00) \\
\midrule
2 & \textbf{Summary:} Person A and Person B are a married couple on a beach at night waiting for grunion. They argue about unmet expectations in their marriage and past disappointments but then reassure each other of their commitment. They share a quiet, intimate moment, and decide to stay together in that spot, hoping the fish might come. 
\newline \textbf{Two prior turns:}
\newline Person A: Or not. \newline Person B: Or not. & Person A: Hmm. & gemma3:4b: (2.00, 4.00, 3.00) \newline gemma3:27b: (2.00, 2.00, 2.00) \newline gpt-oss:20b: (2.00, 2.00, 2.00) \newline gpt-oss:120b: (2.00, 2.00, 2.00) \newline llama3.2:3b: (3.00, 3.00, 4.00) \newline llama3.3:70b: (3.00, 2.00, 3.00) & gemma3:4b: (2.00, 4.00, 3.00) \newline gemma3:27b: (3.00, 2.00, 3.00) \newline gpt-oss:20b: (2.00, 2.00, 2.00) \newline gpt-oss:120b: (3.00, 2.00, 3.00) \newline llama3.2:3b: (3.00, 2.00, 3.00) \newline llama3.3:70b: (3.00, 2.00, 3.00) & (5.00, 4.00, 0.50) \\
\midrule
3 & \textbf{Summary:} Person A and Person B are a married couple on a beach waiting to see fish that never seem to appear. They argue about disappointment and unmet expectations in their lives and relationship. They then gradually reassure each other that despite feeling let down by how things have turned out, they still want to be together. 
\newline \textbf{Two prior turns:}
\newline Person B: It's not champagne. \newline Person A: I guess we don't need glasses. & Person B: Are you cold, huh? Do you want to go home? & gemma3:4b: (2.00, 4.00, 3.00) \newline gemma3:27b: (3.00, 2.00, 4.00) \newline gpt-oss:20b: (4.00, 2.00, 3.00) \newline gpt-oss:120b: (4.00, 2.00, 3.00) \newline llama3.2:3b: (2.00, 2.00, 2.00) \newline llama3.3:70b: (3.00, 2.00, 4.00) & gemma3:4b: (2.00, 4.00, 3.00) \newline gemma3:27b: (3.00, 2.00, 4.00) \newline gpt-oss:20b: (4.00, 2.00, 3.00) \newline gpt-oss:120b: (3.00, 3.00, 3.00) \newline llama3.2:3b: (2.00, 2.00, 2.00) \newline llama3.3:70b: (3.00, 2.00, 4.00) & (4.50, 4.50, 1.00) \\
\midrule
4 & \textbf{Summary:} Person A and Person B are arguing on a beach because Person B is excited about seeing grunion fish while Person A is frustrated and dismisses it as PR hype. Person A insists the outing isn't really about the fish, and complains that they never actually see them. 
\newline \textbf{Two prior turns:}
\newline Person B: We'll see them this year.
\newline Person A: No, we won't. It's pointless. It's like waiting up to see Santa Clause. & Person A: God I feel as though my entire life is going to be spent standing on the beach with my eyes wide open and my hands clasped expectantly waiting for the fish to show up. & gemma3:4b: (2.00, 4.00, 3.00) \newline gemma3:27b: (2.00, 3.00, 2.00) \newline gpt-oss:20b: (2.00, 4.00, 1.00) \newline gpt-oss:120b: (1.00, 3.00, 1.00) \newline llama3.2:3b: (4.00, 5.00, 2.00) \newline llama3.3:70b: (2.00, 4.00, 1.00) & gemma3:4b: (2.00, 4.00, 3.00) \newline gemma3:27b: (2.00, 3.00, 2.00) \newline gpt-oss:20b: (1.00, 3.00, 2.00) \newline gpt-oss:120b: (2.00, 4.00, 2.00) \newline llama3.2:3b: (4.00, 5.00, 2.00) \newline llama3.3:70b: (2.00, 4.00, 1.00) & (1.00, 5.00, 5.00) \\
\midrule
5 & This is the first turn in the conversation; there are no previous turns. & Person A: Look at this, goose bumps . & gemma3:4b: (4.00, 3.00, 4.00) \newline gemma3:27b: (4.00, 4.00, 3.00) \newline gpt-oss:20b: (4.00, 5.00, 3.00) \newline gpt-oss:120b: (4.00, 5.00, 3.00) \newline llama3.2:3b: (5.00, 5.00, 5.00) \newline llama3.3:70b: (5.00, 5.00, 4.00) & gemma3:4b: (4.00, 3.00, 4.00) \newline gemma3:27b: (4.00, 4.00, 3.00) \newline gpt-oss:20b: (4.00, 5.00, 3.00) \newline gpt-oss:120b: (4.00, 5.00, 3.00) \newline llama3.2:3b: (5.00, 5.00, 5.00) \newline llama3.3:70b: (5.00, 5.00, 4.00) & (3.00, 1.50, 2.50) \\
\bottomrule
\end{tabular}
}
\caption{Turns where \textbf{No Reasoner} (hist) and \textbf{EmoLASP} (hist) predictions both have high error.}
\label{tab:high_error_all_models}
\end{table*}

Table~\ref{tab:high_error_all_models} shows the five highest-error cases by MAE where all LLMs fail to predict accurate VAD scores even after applying ASP. In this analysis, we focus on LLMs rather than PLMs, because it is unfair to compare LLMs directly with PLMs that have been fine-tuned on the dataset.

\end{document}